\documentclass[runningheads]{llncs}

\usepackage[T1]{fontenc}
\usepackage{graphicx}
\usepackage{booktabs}
\usepackage{multirow}
\usepackage{amsmath}
\usepackage{amssymb}
\usepackage{url}
\usepackage{hyperref}
\usepackage{placeins}
\hypersetup{hidelinks}
\graphicspath{{figures/}}

\newcommand{\ehr}{EHR}
\newcommand{\cxr}{CXR}
\newcommand{\ecg}{ECG}
\newcommand{\FedAvg}{\textsc{FedAvg}}
\newcommand{\FedCoRe}{\textsc{FedCoRe}}
\newcommand{\NoCompletion}{\textsc{NoCompletion}}

\newcommand{\auroc}{AUROC}

\begin{document}

\title{\FedCoRe{}: Target-Adaptive Completion for Missing Modalities in Healthcare Federated Learning}
\titlerunning{Target-Adaptive Completion for Missing Modalities}

\author{Holger R. Roth\thanks{Corresponding author.} \and
Ziyue Xu \and
Peter Cnudde}
\authorrunning{H. R. Roth et al.}
\institute{NVIDIA, Santa Clara, CA, USA\\
\email{hroth@nvidia.com}}

\maketitle

\begin{abstract}
Federated multimodal models often assume every site has every modality, although hospitals differ in access to EHRs, chest radiographs, and ECGs.
We study this setting on a MIMIC-derived respiratory deterioration task with simulated FL clients and introduce \FedCoRe{} (Federated Cross-Modal Representation Completion).
\FedCoRe{} learns representation- or logit-space corrections rather than generating synthetic ECGs or CXR images.
When a client observes a modality that may be missing at deployment, it evaluates the same example with and without that modality to obtain paired supervision.
Only clients with such pairs update the completion module, and validation may retain the unchanged prediction.
We freeze the trained multimodal predictor during evaluation so that measured differences come only from completion.
Hiding ECG reduced AUROC by about 0.085; paired-example FedAvg restored 0.0415 AUROC, or 49.0\% of the lost performance.
We therefore report two distinct effects: paired-example \FedAvg{} partially recovers the missing-\ecg{} gap, while validation-selected completion is a task-specific classifier-logit correction rather than literal \ecg{} recovery.
For CXR, effect-aware completion recovers 52.8\% of the loss in a controlled test where CXR is hidden.
Paired-example \FedAvg{} transfers part of this effect, but validation keeps the no-completion baseline for deployment cases whose inputs lack CXR.
Thus, \FedCoRe{} should be read as a validation-gated completion/correction framework: it can recover missing-modality signal in supported settings, but it should be deployed only when paired examples and validation evidence support that modality.
\end{abstract}

\section{Introduction}
\begin{figure}[htbp]
  \centering
  \includegraphics[width=0.84\textwidth]{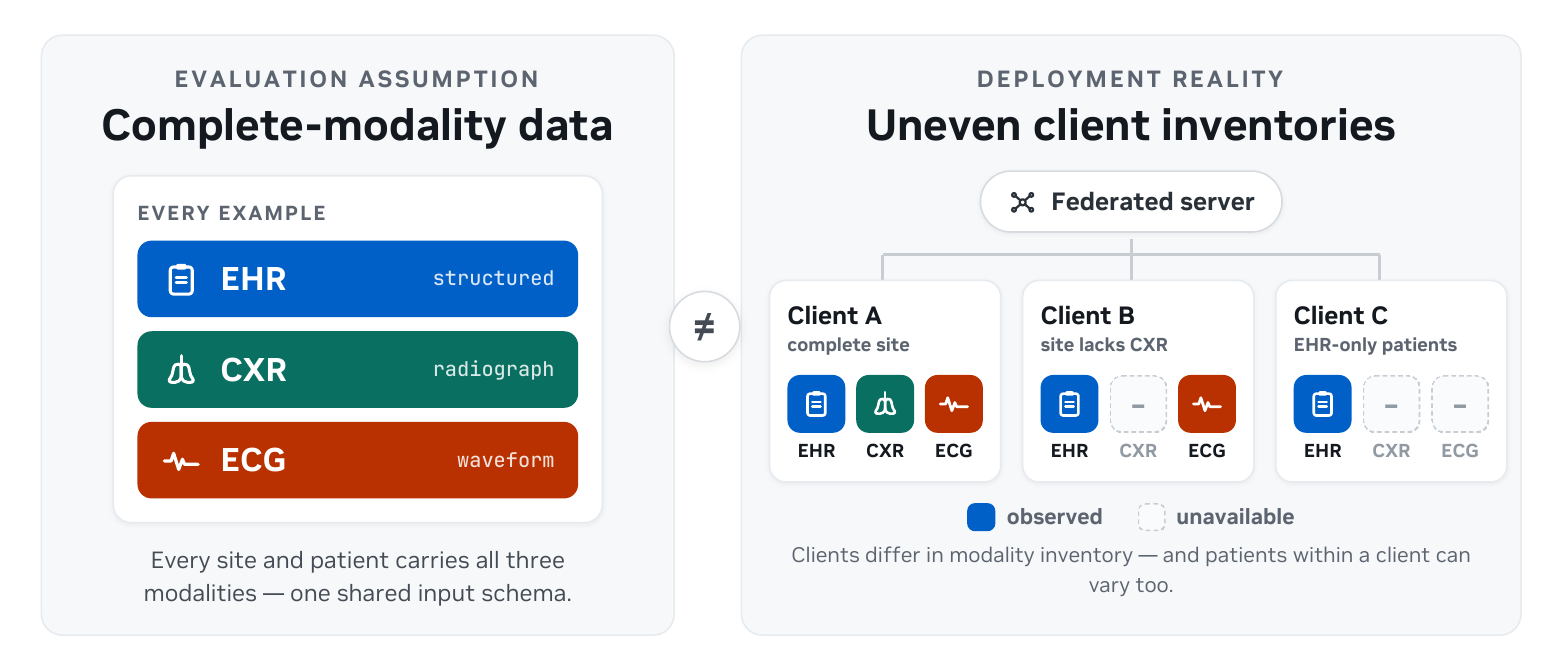}
  \caption{Complete-modality evaluation contrasts with federated deployment, where sites and patients may have different subsets of EHR, CXR, and ECG.}
  \label{fig:modality-gap}
\end{figure}

Multimodal clinical prediction increasingly combines structured context, radiology images, physiological signals, and text.
Complete-modality evaluation is often overly optimistic for federated learning (FL): some hospitals may support different modalities, while others may include patients with only a subset of the observed modalities.
Figure~\ref{fig:modality-gap} illustrates this gap between evaluation and deployment.

Our question is whether clients observing a modality can teach a lightweight correction that benefits clients or patients missing it, without centralizing raw records~\cite{rieke2020future}.
Here, ``completion'' means adding a learned residual in representation or classifier-logit space, not generating a synthetic ECG or CXR.
We separate two possible sources of improvement: \textbf{(1)} robust predictor training and \textbf{(2)} the completion operator itself.
Accordingly, we ask whether collaborative training through FL helps missing-modality patients and, with that predictor frozen, whether completion adds further signal.

\FedCoRe{} learns task-aware corrections, updates each operator only from clients with valid modality pairs, aggregates those updates, and uses validation to apply the operator or retain the unchanged prediction.

\textbf{Related Work.} \FedAvg{} is the canonical decentralized optimization algorithm~\cite{mcmahan2017communication}; platforms such as Flower~\cite{beutel2020flower} or NVIDIA FLARE~\cite{roth2022nvflare} can coordinate healthcare FL while keeping raw records at their source.
Recent reviews identify modality heterogeneity as a central barrier~\cite{thrasher2025mmflreview} and prior multimodal FL addresses missingness through pseudo-modality generation~\cite{yan2025fedpmg}, cross-contrast synthesis~\cite{wang2023fedmedgan,dalmaz2024pflsynth}, retrieval~\cite{poudel2024carmfl}, contrastive ensembles~\cite{yu2023creamfl}, embedding transfer~\cite{le2025fedmekt}, and prototype/mask completion~\cite{bao2023prototype}. \FedCoRe{} instead learns modality-specific corrections from paired outputs without requiring universal modality availability.
Raw records remain local, but we do not claim formal privacy guarantees; model updates may leak information without mechanisms such as secure aggregation or differential privacy~\cite{kairouz2021advances}.


\section{Method}

\textbf{Intuition.}
Suppose one client has \ehr{}, \cxr{}, and \ecg{}, while another lacks \ecg{}. The first client compares predictions with and without \ecg{} to train an ECG-specific completion operator. The server aggregates operator updates only from clients that can form such pairs; the second client applies the operator only when validation shows that it helps. Thus, \FedCoRe{} creates paired views, trains correction operators, aggregates only valid updates, and validates deployment.

\textbf{Paired views.}
Let client $k$ contain examples $\mathcal{D}_k=\{(x_i^{M_i},y_i)\}_{i=1}^{n_k}$, where $y_i$ is the clinical outcome and $M_i \subseteq \mathcal{M}$ is the observed modality set; $\mathcal{M}=\{\mathrm{EHR},\mathrm{CXR},\mathrm{ECG}\}$.
A fixed encoder $f_\phi$ and prediction head $g_\psi$ map the modalities available for an example to an internal representation $h_i(M_i)$ and classifier logits $\ell_i(M_i)$:
\begin{equation}
    h_i(M_i)=f_\phi(x_i^{M_i}), \qquad \ell_i(M_i)=g_\psi(h_i(M_i)).
\end{equation}
Let $t$ denote the input modality we want to complete when it is missing at deployment.
Paired supervision exists only if $t\in M_i$ and $M_i\setminus\{t\}\neq\emptyset$.
We evaluate each such example twice to form a pair:
\begin{equation}
    h_i^{+t} = h_i(M_i), \qquad
    h_i^{-t} = h_i(M_i \setminus \{t\}),
\end{equation}
Here, $h_i^{+t}$ is the representation with $t$ observed and $h_i^{-t}$ is the representation after removing $t$.
The corrected output introduced below is the \emph{completed} representation or logit.
Examples without $t$, or with only $t$, cannot supervise its completion.

\textbf{Completion operator.}
\FedCoRe{} learns the change induced by modality $t$.
For pooled hidden states, a modality-specific operator $C_{\theta_t}$ predicts a correction to the $t$-removed representation:
\begin{equation}\label{eq:representation-completion}
    \tilde{h}_i^t = h_i^{-t} + \alpha_t C_{\theta_t}(h_i^{-t}),
\end{equation}
The scalar $\alpha_t$ controls how strongly that correction is applied; a zero value leaves the original representation unchanged.
For logit-delta completion, the operator instead corrects the frozen classifier output directly:
\begin{equation}\label{eq:logit-completion}
    \tilde{\ell}_i^t = \ell_i^{-t} + \alpha_t C_{\theta_t}(h_i^{-t}).
\end{equation}
Thus, $C_{\theta_t}$ can operate either in representation space or in classifier-logit space: it predicts a hidden-state residual in Eq.~\eqref{eq:representation-completion}, or an additive logit residual in Eq.~\eqref{eq:logit-completion}.
In both cases, the paired $t$-present pass provides the training reference: $h_i^{+t}$ for representation completion and $\ell_i(M_i)$ for logit completion.
The formulation is backbone-agnostic: it requires only paired $t$-present and $t$-removed representations or logits.

For CXR, we use an effect-aware logit interface. Let
\[
d_i^{\mathrm{CXR}}=\ell_i(M_i)-\ell_i(M_i\setminus\{\mathrm{CXR}\}),
\qquad
\Delta_i^{\mathrm{CXR}}=\gamma_i r_{\theta}(h_i^{-\mathrm{CXR}}),
\]
where \(d_i^{\mathrm{CXR}}\) is the observed two-logit CXR effect and
\(\gamma_i\in[0,1]\) is an instance gate. Deployment uses
\[
\tilde \ell_i^{\mathrm{CXR}}
=
\ell_i^{-\mathrm{CXR}}+\alpha_{\mathrm{CXR}}\Delta_i^{\mathrm{CXR}}.
\]
The CXR loss adds Smooth-L1 effect matching, probability consistency, gate supervision, and gate sparsity to the task/logit terms. The implementation uses \(\lambda_d\), \(\lambda_p\), \(\lambda_g\), and
\(\lambda_s\) for effect matching, probability consistency, gate supervision,
and gate sparsity, respectively.

\textbf{Local objective.}
Each client learns the correction from its valid pairs using a task-aware objective:
\begin{equation}
\mathcal{L}_i^t =
\lambda_{\mathrm{task}} \mathrm{CE}(g_\psi(\tilde{h}_i^t), y_i)
+ \lambda_{\mathrm{logit}} \lVert g_\psi(\tilde{h}_i^t)-g_\psi(h_i^{+t})\rVert_2^2
+ \lambda_{\mathrm{rep}} \lVert \tilde{h}_i^t-h_i^{+t}\rVert_2^2 .
\end{equation}
The three terms preserve label prediction, align the completed and $t$-present logits, and regularize the completed representation, respectively.
The main missing-\ecg{} result retains the task and logit terms and sets $\lambda_{\mathrm{rep}}=0$ after validation. 

\textbf{Aggregation by available supervision.}
Clients share completion modules without assuming that every client can supervise every modality.
For modality $t$, the valid paired set at client $k$ is
\begin{equation}
    \mathcal{P}_{k,t}=\{i \in \mathcal{D}_k : t \in M_i,\; M_i \setminus \{t\}\neq \emptyset\}
\end{equation}
Clients with $|\mathcal{P}_{k,t}|=0$ return no completion weights and receive zero aggregation weight.
The server averages the remaining updates in proportion to the number of valid pairs:
\begin{equation}
    \theta_t^{r+1} =
    \sum_{k=1}^{K}
    \frac{|\mathcal{P}_{k,t}|}{\sum_j |\mathcal{P}_{j,t}|}
    \theta_{k,t}^{r+1}.
\end{equation}
Consequently, a client that cannot supervise completion for $t$ cannot dilute or corrupt its global completion operator.

\begin{figure}[htbp]
\centering
\includegraphics[width=0.94\linewidth]{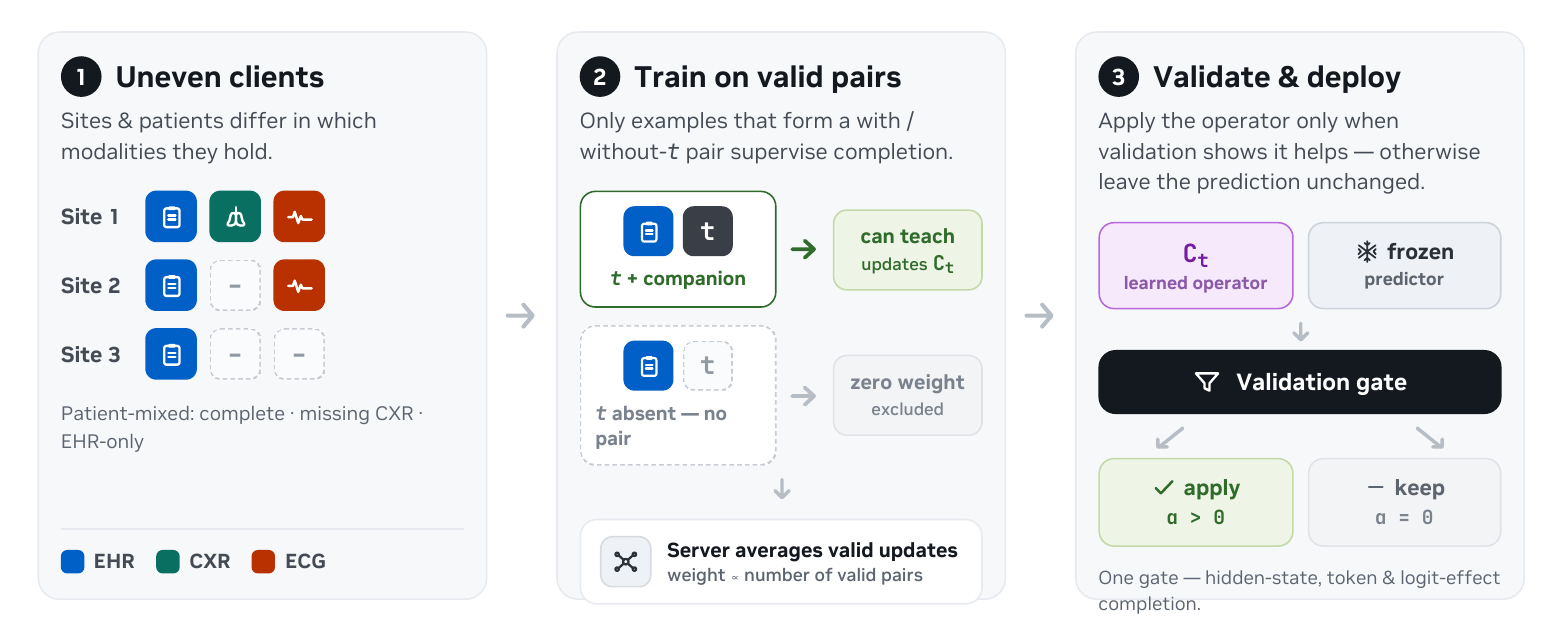}
\caption{\FedCoRe{} workflow: paired observations train the completion operator, sites without valid pairs receive zero weight, and validation applies completion or leaves predictions unchanged.}
\label{fig:method}
\end{figure}

\textbf{Validation-based deployment.}
Finally, validation selects a candidate $c=(s,\alpha,\tau)\in\mathcal{C}_t$: an operator source $s$, completion strength $\alpha$, and optional gate threshold $\tau$.
The no-completion candidate $c_0$ has $\alpha=0$ and leaves predictions unchanged.
For deployment, $V_{\mathrm{miss}}^t(c)$ is computed on cases whose observed inputs exclude $t$; for a frozen-predictor completion test, it uses complete cases after deliberately removing $t$.
$V_{\mathrm{safe}}(c)$ optionally measures a protected validation view, such as complete-modality cases.
We select
\begin{equation}
    c_t^\star = \arg\max_{c \in \mathcal{C}_t} V_{\mathrm{miss}}^t(c)
    \quad \mathrm{s.t.}\quad
    V_{\mathrm{miss}}^t(c)\ge V_{\mathrm{miss}}^t(c_0),\;
    V_{\mathrm{safe}}(c)\ge V_{\mathrm{safe}}(c_0)-\epsilon .
\end{equation}
The second constraint is omitted when no separate safety view is used.
If no completion candidate qualifies, $c_0$ leaves the original prediction unchanged.
Figure~\ref{fig:method} summarizes the complete workflow.

\FloatBarrier


\section{Task and Federated Setup}

\textbf{Cohort and label.} We construct a MIMIC-derived benchmark using public datasets from \mbox{PhysioNet}\footnote{\url{https://physionet.org/}}~\cite{goldberger2000physionet}: MIMIC-IV~\cite{physionetMimicIv31,johnson2023mimiciv}, MIMIC-CXR~\cite{physionetMimicCxr200,johnson2019mimiccxr}, MIMIC-CXR-JPG~\cite{physionetMimicCxrJpg210,johnson2019mimiccxrjpg}, and MIMIC-IV-ECG~\cite{gow2023mimicivecg}. The benchmark links structured \ehr{} context, a \cxr{} image, and a diagnostic \ecg{} record. The task is binary respiratory deterioration within 48 hours of the later \ecg{}--\cxr{} ``index'' timestamp, operationalized as future invasive ventilation, ICU transfer, or all-cause in-hospital death. 
To reduce leakage, structured context is pre-index, outcome event times are strictly post-index, ICU-transfer labels use post-index transfer times, and encounters in which the patient was already in an ICU at the index time are excluded. The ventilation endpoint counts only new post-index invasive ventilation events. \ecg{} and \cxr{} observations are matched within 6 hours. The cohort contains 13,914 encounters from 11,331 patients, split patient-disjoint into 9,777 train, 2,056 validation, and 2,081 test examples; the positive class is 304/13,914 (2.2\%).

\textbf{FL modality heterogeneity.} Let $M_i \subseteq \{\mathrm{EHR},\mathrm{CXR},\mathrm{ECG}\}$ be the observed modalities for example $i$. We simulate client-level missingness and patient-mixed missingness, where a client contains both complete and incomplete examples. Splits are patient-disjoint across train/validation/test and FL clients. Table~\ref{tab:templates} summarizes the main templates used for the missing-\ecg{} frozen-predictor completion test.

\begin{table}[htbp]
\centering
\caption{Core FL heterogeneity templates for the missing-\ecg{} frozen-predictor completion test. Missing-\ecg{} mass is expected test-set patient mass under the client mask; paired sources are examples in which ECG is observed and can supervise completion.}
\label{tab:templates}
\scriptsize
\resizebox{\linewidth}{!}{%
\begin{tabular}{lccccl}
\toprule
Template & Clients & Full-modality clients & Missing-\ecg{} mass & Paired-example structure & Purpose \\
\midrule
No full-modality client & 4 & 0 & 66.3\% & ECG-present but no complete site & Negative control \\
Patient-mixed any-missing & 4 & Mixed & 30.0\% & Complete and ECG-present patients within clients & Main heterogeneous setting \\
Patient-mixed heavy & 4 & Mixed & 36.3\% & Heavier missing-ECG patient mix & Stress setting \\
\bottomrule
\end{tabular}
}
\end{table}

\section{Experiments}
\textbf{Baselines and controls.}
We compare no completion, local-only completion, paired-example \FedAvg{} completion, and validation-selected completion; full-modality runs provide a reference, not a centralized pooled or directly reimplemented prior-method comparison.
CXR controls include server-selected and model-soup sources, scalar logit recalibration, pseudo-CXR tokens, and forced fixed-scale completion.
In Table~\ref{tab:ecg}, ``Before'' is \NoCompletion{}, \FedAvg{} uses the aggregated operator with a validation-selected strength, and ``Val-selected'' additionally chooses among local and \FedAvg{} sources.
FedCoRe operates on previously trained global multimodal predictors, independent of whether predictor training is centralized or federated.
During controlled completion experiments, each predictor is held fixed so that performance changes are attributable to the target-specific completion operators trained and aggregated across clients.
The ECG predictor was obtained through prior modality-heterogeneous federated training, whereas the CXR reference predictors were trained separately on the complete EHR+CXR training view.

\textbf{CXR configuration.}
Missing-\cxr{} experiments use a frozen Qwen3-VL-8B-Instruct predictor with LoRA rank 8 adapters, pre-index \ehr{} context, and \cxr{} images.
The effect-aware gated MLP uses hidden size 4096, dropout 0.25, learning rate $5{\times}10^{-5}$, 80 local operator-training epochs, batch size 128, uncapped completion train/validation sets, and balanced task weighting.
Its loss weights are $\lambda_{\mathrm{task}}=1$, $\lambda_{\mathrm{logit}}=8$, and $\lambda_{\mathrm{rep}}=0$, with Smooth-L1 effect matching, $\lambda_p=0.5$, $\lambda_g=1.0$, and $\lambda_s=0.005$.
Validation considers $\alpha\in\{0,.1,.25,.5,.75,\allowbreak 1,1.25,1.5,\allowbreak 2,3,4,6,8\}$ and gate thresholds $\{0,.1,.25,.5,.75\}$.

\textbf{ECG/EHR configuration.}
Supporting \ecg{}/\ehr{} rows use an earlier Qwen-family backend with lead-II ECG patch tokens.
The main \ecg{} operator is a two-layer logit-delta MLP with hidden size 2048, dropout 0.3, learning rate $10^{-4}$, batch size 128, 50 local epochs, and one \FedAvg{} completion round across four simulated clients, with $\lambda_{\mathrm{task}}=1$, $\lambda_{\mathrm{logit}}=4$, and $\lambda_{\mathrm{rep}}=0$.

\textbf{Selection protocol.}
Candidate completion operators are selected using validation data only, and every strength grid includes $\alpha=0$ (no completion).
For \ecg{}, each source selects $\alpha\in\{0,2,4,8,12\}$ by missing-ECG validation \auroc{} subject to nonnegative gain; ``Val-selected'' then chooses the source with the greatest validation gain.
The no-full-modality, any-missing, and heavy settings evaluate 3, 11, and 8 sources, respectively.
All positive reported rows have unique validation maxima; any tie is resolved by the smallest $|\alpha|$ and a deterministic source identifier.
For \cxr{}, we evaluate the fixed operator over the 13 strengths and five gate thresholds above.
A CXR candidate must improve remove-CXR validation \auroc{}, have at least 0.02 full-minus-removed validation headroom, and reduce full-view validation \auroc{} by no more than $\epsilon=0.005$.
The validation/test manifests contain 2,056/2,081 examples with 44/42 positives.
For the ECG-present reference, the same frozen predictor receives \cxr{}, \ehr{}, and \ecg{}; the missing view removes only ECG on the same manifest.
Table~\ref{tab:ecg} changes the available completion sources and paired-example pattern, not the frozen predictor.
Test predictions were saved during replay but summarized only after validation fixed the source, strength, gate threshold, or no-completion choice; test metrics were never used for selection.

\textbf{Metrics.}
Our primary endpoint is the absolute change in \auroc{} among cases missing the modality to be completed.
Secondary measures are relative lift and, when available, aggregate deployment \auroc{} change.
We report positive-row counts and selected strength; main \ecg{}/\cxr{} analyses also include AUPRC and patient-level paired bootstrap 95\% CIs.
The intervals condition on the validation-selected operator and exclude source, strength, and gate selection uncertainty.
With 2.2\% prevalence and 42 test positives, \auroc{} does not establish precision or clinical-threshold utility; results are model-development evidence.

\section{Results \& Discussion}
\begin{figure}[htbp]
\centering
\includegraphics[width=0.96\linewidth]{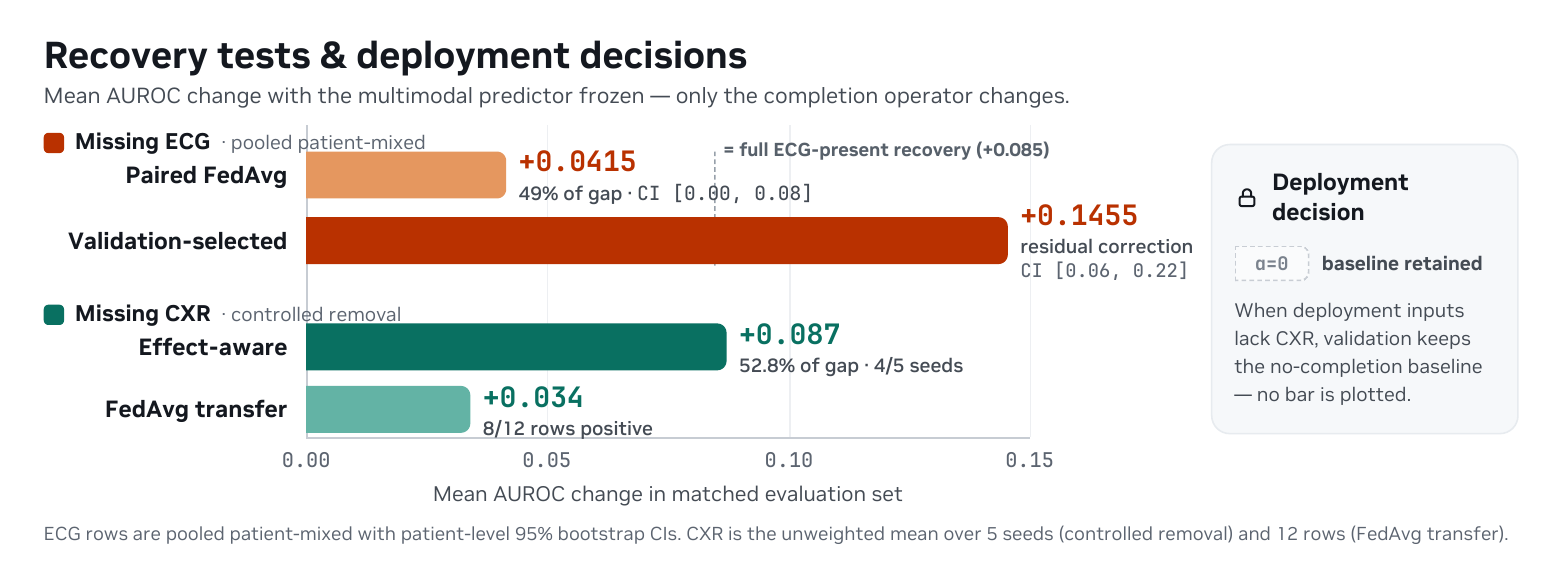}
\caption{Observed gains with the multimodal predictor frozen. Bars report matched target-removal/recovery tests rather than deployment gains; the box separately reports CXR deployment, where validation selected $\alpha=0$ and retained the baseline.}
\label{fig:recovery}
\end{figure}

\suppressfloats[t]
\textbf{ECG is the most reliably completed modality.}
Across the exploratory matrix, missing \ecg{} improves in 69/70 rows, with mean +0.059 \auroc{} and +9.2\% relative lift.
Missing \ehr{} improves more modestly in 20/20 rows (mean +0.027), while missing \cxr{} depends more strongly on the backbone and completion interface.
Because the modalities use different source inputs, backbones, and completion interfaces, these numbers are not a ranking of clinical importance.
Figure~\ref{fig:recovery} summarizes the main decisions, with detailed \ecg{} and \cxr{} results in Tables~\ref{tab:ecg} and~\ref{tab:cxr_qwen3vl}.

\textbf{Paired-example \FedAvg{} recovers about half of the missing-ECG gap.}
With ECG present, the frozen predictor reaches 0.6627 \auroc{}; removing ECG reduces it to 0.5781.
Across the six patient-mixed rows, \FedAvg{} raises \auroc{} to 0.6195, adding +0.0415 [0.0023, 0.0778] and recovering 49.0\% of the lost performance.
The predictor and patients are unchanged; only the completion operator differs.
In the no-full-modality-client control, validation leaves the prediction unchanged because none of the available corrections helps.

\textbf{Validation-selected ECG completion acts as residual correction.}
It reaches 0.7235 \auroc{}, an improvement of +0.1455 [0.0617, 0.2213] that exceeds the ECG-present reference.
We therefore interpret this result as classifier-logit correction beyond modality recovery, not literal ECG reconstruction.
AUPRC changes remain unstable with 42 positive test examples; selected-source validation/test deltas are +0.120--0.136/+0.104--0.169.

\begin{table}[htbp]
\centering
\caption{Frozen-predictor missing-\ecg{} test. Both methods select scale on validation; ``Val-selected'' also selects source. CIs use shared patient-level resamples across the six matched patient-mixed predictions. Full ref is ECG-present replay; reference-normalized gain is $(\mathrm{After}-\mathrm{Before})/(\mathrm{Full~ref}-\mathrm{Before})$, where values above 100\% denote correction beyond the reference.}
\label{tab:ecg}
\scriptsize
\resizebox{\linewidth}{!}{%
\begin{tabular}{lllccccccc}
\toprule
Scenario & Completion & Seeds & Before & After & Full ref & Ref.-norm. & $\Delta$ \auroc{} [95\% CI] & AUPRC & Positive \\
\midrule
No full-modality client & \FedAvg{} & 7,17,42 & 0.5781 & 0.5781 & 0.6627 & 0.0\% & +0.0000 [0.0000, 0.0000] & 0.0375 / 0.0375 & 0/3 \\
No full-modality client & Val-selected & 7,17,42 & 0.5781 & 0.5781 & 0.6627 & 0.0\% & +0.0000 [0.0000, 0.0000] & 0.0375 / 0.0375 & 0/3 \\
\textbf{Patient-mixed combined} & \textbf{\FedAvg{}} & \textbf{6 rows} & \textbf{0.5781} & \textbf{0.6195} & \textbf{0.6627} & \textbf{49.0\%} & \textbf{+0.0415 [0.0023, 0.0778]} & \textbf{0.0375 / 0.0368} & \textbf{6/6} \\
\textbf{Patient-mixed combined} & \textbf{Val-selected} & \textbf{6 rows} & \textbf{0.5781} & \textbf{0.7235} & \textbf{0.6627} & \textbf{171.8\%} & \textbf{+0.1455 [0.0617, 0.2213]} & \textbf{0.0375 / 0.0495} & \textbf{6/6} \\
\bottomrule
\end{tabular}
}
\end{table}

\textbf{Additional modality tests.} A single missing-\ecg{} scenario gives +0.095 mean \auroc{} for \FedAvg{} and +0.116 for validation-selected completion across six seeds; missing-\ehr{} is weaker but non-harmful (+0.014, three seeds).

\textbf{CXR is partially recoverable when removed.}
With Qwen3-VL, we evaluate complete cases with CXR present, deliberately remove CXR, and apply the effect-aware logit operator.
Across five seeds, completion adds +0.087 \auroc{} on average (4/5 positive) and recovers 52.8\% of the full-vs-missing gap.
The effect varies: seed 23 leaves the prediction unchanged, and only two per-seed intervals exclude zero; we therefore report paired CIs per seed rather than treat seeds as independent clinical replications.
AUPRC improves by +0.010 on average but is unstable with 42 positives.
Scalar recalibration gives zero rank-metric lift.
Pseudo-\cxr{} tokens are a negative ablation: validation rejects them, while forced completion reduces \auroc{} by 0.082.
This is evidence that some CXR-induced logit signal is recoverable from the remaining inputs, not that deployment without CXR is solved.

\begin{table}[htbp]
\centering
\caption{Frozen-predictor missing-\cxr{} completion test. Validation selects scale and gate threshold; only scale is shown. Delta intervals are paired patient-level bootstrap 95\% CIs. Means and gap recovery are unweighted across seeds.}
\label{tab:cxr_qwen3vl}
\scriptsize
\resizebox{\linewidth}{!}{%
\begin{tabular}{lccccccc}
\toprule
Seed & Full \auroc{} & Missing & Completed & $\Delta$ \auroc{} [95\% CI] & Gap rec. & AUPRC before/after & $\alpha$ \\
\midrule
7 & 0.6673 & 0.5237 & 0.6454 & +0.1217 [0.0005, 0.2330] & 84.7\% & 0.0204 / 0.0409 & 0.50 \\
17 & 0.6969 & 0.5933 & 0.6228 & +0.0295 [-0.0520, 0.1104] & 28.5\% & 0.0402 / 0.0331 & 0.75 \\
23 & 0.7303 & 0.6187 & 0.6187 & +0.0000 [0.0000, 0.0000] & 0.0\% & 0.0297 / 0.0297 & 0.00 \\
31 & 0.7072 & 0.5188 & 0.6594 & +0.1405 [0.0301, 0.2464] & 74.6\% & 0.0255 / 0.0345 & 0.10 \\
42 & 0.7192 & 0.5307 & 0.6739 & +0.1431 [-0.0037, 0.2826] & 75.9\% & 0.0327 / 0.0611 & 0.10 \\
\midrule
Mean & 0.7042 & 0.5570 & 0.6440 & +0.0870 & 52.8\% & 0.0297 / 0.0399 & -- \\
\bottomrule
\end{tabular}
}
\end{table}

\textbf{CXR completion transfers through \FedAvg{} but is not deployed when unsupported.}
Paired-example \FedAvg{} improves the remove-CXR test in 8/12 rows (mean +0.034 \auroc{}); server-selected sources are weaker (+0.017, 2/6 positive), and source-subset operators are negative (0/12).
For deployment cases whose observed inputs lack CXR, validation selects $\alpha=0$.
\FedCoRe{} therefore retains the no-completion baseline instead of applying an unsupported CXR correction.


\section{Conclusion}
\FedCoRe{} learns modality-specific operators from valid pairs, excludes uninformative clients from aggregation, and can retain the unchanged prediction. Missing-\ecg{} benefits from completion; missing-\cxr{} shows recoverable signal in remove-CXR tests but not unsupported deployment cases. Controlled four-client masks impose modality availability rather than reflecting natural missingness or institutional covariate shift; external multi-site validation remains future work. 


\textbf{Code and AI use disclosure.}
Experiments used NVIDIA FLARE\footnote{\url{https://github.com/NVIDIA/NVFlare}}. Codex and ChatGPT assisted with experimentation and writing, and Claude Design with figures; the authors verified all content and remain responsible.

\subsubsection{Disclosure of Interests.}
The authors have no competing interests to declare that are relevant to the content of this article.

\bibliographystyle{splncs04}
\bibliography{references}

\end{document}